\documentclass{article} 
\usepackage{nips15submit_e,times}
\usepackage{hyperref,amsmath,graphicx,amsfonts}
\usepackage{url}

\title{Learning a Driving Simulator}

\author{
Eder Santana
\thanks{This work was done during a Summer 2016 internship at \texttt{comma.ai}. The companion material of this paper
can be found online: http://research.comma.ai} \\
University of Florida\\
eder@comma.ai
\And
George Hotz \\
\texttt{comma.ai} \\
george@comma.ai
}

\nipsfinalcopy 

\begin{document}

\maketitle

\begin{abstract}
\texttt{comma.ai}'s approach to Artificial Intelligence for self-driving cars\footnote{https://www.youtube.com/watch?v=KTrgRYa2wbI} is based on an agent that learns to clone driver behaviors and plans maneuvers by simulating future events in the road.
This paper illustrates one of our research approaches for driving simulation. One where
we learn to simulate. Here we investigate variational autoencoders with classical and learned cost functions
using generative adversarial networks for embedding road frames. Afterwards, we learn a transition model
in the embedded space using action conditioned Recurrent Neural Networks. We show that our approach can keep predicting
realistic looking video for several frames despite the transition model being optimized without a cost function in the pixel space.
\end{abstract}

\section{Introduction}
\label{sec:intro}

Self-driving cars are one of the most promising prospects for near term Artificial Intelligence research.
The state-of-the-art of this research leverages large amounts of labeled and contextually rich data,
which abounds in driving. From a complex perception and controls perspective, the technology to correctly solve driving
can potentially be extended to other interesting tasks such as action recognition from videos and planning.
An approach for self-driving cars that is economically attractive while still expanding the AI frontier
should be based on vision, which is also the main sensor used by a human driver.

Vision based controls and reinforcement learning had recent success in the literature \cite{schimidhubercar} \cite{atarinature} \cite{alphago} \cite{googlerobotarm}
mostly due to (deep, recurrent) neural networks and unbounded access to world or game iteraction. Such interactions
provide the possibility to revisit states with new policies and to simulate future events for training deep neural network based
controllers. For example, Alpha Go \cite{alphago} used deep convolutional neural networks to predict the probability of winning the game
from a given state using the outcomes of several games, lots of which were playing but previous iterations of Alpha Go itself. The Go game engine was also
used for simulating possible unfoldings of the game and Markov Chain Tree Search. For learning to play Torcs \cite{schimidhubercar} or Atari \cite{atarinature}
using only the game screen, it was also necessary to use several hours of game play.

For the controls problems where unbounded iteraction of a learning agent with the world is not easily affordable there are two possible solutions.
The solutions are to either hand code a simulator or to learn to simulate. Hand coding a simulator involves domain expertise for defining
rules of physics and modelling the randomness of the world. But for most cases, such expertise already encompass all the information for
posing the related controls problem, for example flight simulators \cite{flyandcontrol}, robot walking \cite{bipedal}, etc.

Here instead we focus on learning to simulate aspects of the world from examples of a human agent. We focus on generating video streams of a
front facing camera mounted in the car windshield. Previous successful approaches for learning to simulate for controls were based on
state space representations of the agent physics \cite{nghelicopter}. Other vision only models focused on low dimensional \cite{e2c} or videos with simple
textures, such as those of Atari games \cite{condpred} \cite{learn2think}. Texture rich approaches were based on passive video prediction
for action recognition \cite{beyondmse} without the possibility of an external agent modifying the future of the video and without generating a compact representations
of the data.

This paper contributes to this learned video simulation literature. No graphics engine or world model assumptions are made. We show that it is possible
to train a model to do realistic looking video predictions, while
calculating a low dimensional compact representation and action conditioned transitions. We leave controls on the simulated world for future work.
In the next section we describe the dataset we used to study video prediction of real world road videos.

\section{Dataset}
\label{sec:data}

\begin{figure}[t]
 \centering
 \includegraphics[scale=0.15]{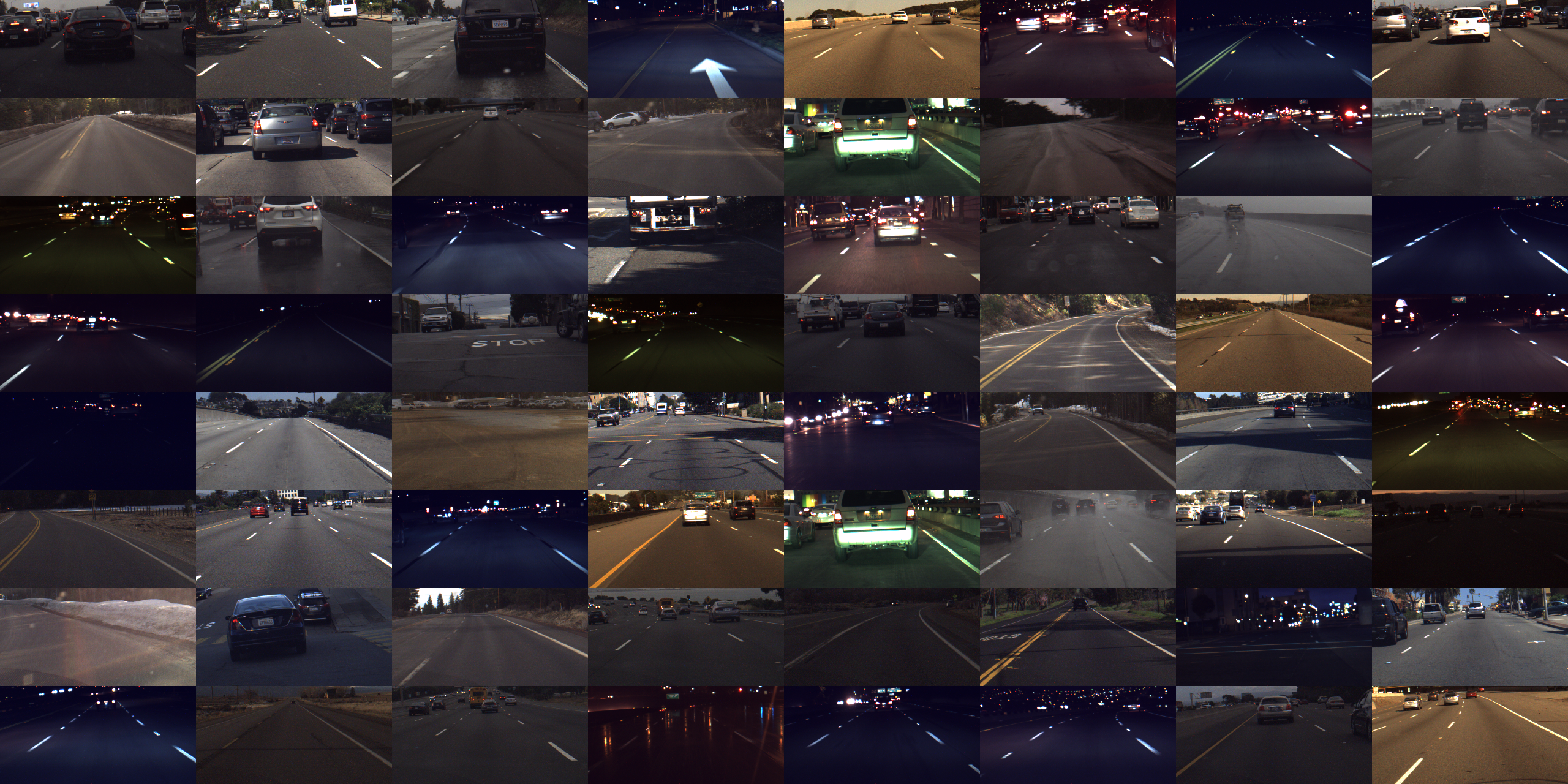}
 \caption{$80 \times 160$ samples from the driving dataset.}
 \label{fig:samples}
\end{figure}

We are publicly releasing part of the driving data used in this paper. The dataset has the same video and sensors
used in the \texttt{comma.ai} self-driving car test platform.

We mounted a Point Grey camera in the windshield of an Acura ILX 2016, capturing pictures of the road at 20 Hz.
In the released dataset there is a total of 7.25 hours of driving data divided in 11 videos. The released video frames are
$160\times 320$ pixels regions from the middle of the captured screen.
Beyond video, the dataset also has several sensors that were measured in different frequencies and interpolated to 100Hz.
Example data coming from sensors are the car speed, steering angle,
GPS, gyroscope, IMU, etc. Further detail about the dataset and measurement equipment can be found online in the companion website.

We also record the time stamps at which these sensors were measured and the time stamps the camera frames were
captured. In our production pipeline the measurement times and simple linear interpolation are used to sync the sensors and camera frames.
We are releasing the raw sensors and camera frames in HDF5 files for easy to use in machine learning and controls software.

In this paper we focus on the camera frames, steering angle and speed.
We preprocessed the camera frames by downsampling them to $ 80\times 160 $ and renormalizing the pixel values
between -1 and 1.
Sample images of this transformed dataset are shown in Fig. \ref{fig:samples}.

In the next section we define the problem investigated in this paper.

\section{Problem definition}
Let $x_t$ denote the $t$-th frame of the dataset. $X_t = \{x_{t-n+1}, x_{t-n+2}, ..., x_{t}\}$ denotes a $n$ frames long video.
These frame are associated with control signals $S_t = \{s_{t-n+1}, s_{t-n+2}, ..., s_{t}\}$, $A_t = \{a_{t-n+1}, a_{t-n+2}, ..., a_{t}\}$
corresponding to the ego speeds and steering angles of the car. Learning to simulate the road can be defined as estimating
the function $F : \mathbb{R}^{80\times 160\times 3 \times n} \times \mathbb{R}^n \times \mathbb{R}^n \rightarrow \mathbb{R}^{80\times 160\times 3}$ that
predicts $x_{t+1} = F(X_t, S_t, A_t)$.

Note that this definition is high dimensional and the dimensions are highly correlated. In Machine Learning problems like those are known to converge slowly
or underfit the data \cite{principebook}. For example, previous work \cite{mydpcn} showed that when using convolutional dynamic networks
without proper regularization, the model simply learns a transformation close to the identity while still getting low cost scores due to correlations in time.
Previous approaches learned the function $F$ directly, but were demonstrated only in very simple, artificial videos \cite{condpred}.
Recently, it was shown \cite{mydpcn} \cite{beyondmse} that it is possible to generate videos with complex textures, but the referred work did not address the problem
of action conditioned transitions and they do not generate a compact intermediate representation of the data. In other words, their models are fully convolutional with
no downsampling and no low dimensional hidden codes. We believe that compact intermediate representations
are important for our research because probabilities, filtering, and consequently controls, are ill defined in very high dimensional dense spaces \cite{probhigh}.

As far as we know, this is the first paper that tries to learn video predictions from real world highway scenes.
As such, in this first paper, we decided to learn the function $F$ in a piecewise manner, so we could debug its parts separately.
First we learned an autoencoder to embed the frames $x_t$ into a Gaussian latent space $z_t \in \mathbb{R}^{2048}$, where the dimensionality
2048 was chosen experimentally and the Gaussian assumption enforced with variational autoencoding Bayes \cite{VAE}. This first step
simplified the problem from learning transitions directly in the pixel space to learning in the latent space. Beyond that, assuming
that an autoencoder is correctly learned respecting the latent space Gaussianity, we can generate realistic looking videos as long as the
transition model never leaves the high density region of the embedding space. This high density region is a hypersphere of radius
$\rho$, which in its turn is function of the dimensionality of embedding space and variance of Gaussian prior. We go into the details of
the autoencoder and the transition model in the next section.

\section{Driving simulator}
\label{sec:raodsim}

\begin{figure}[t]
 \centering
 \includegraphics[scale=0.37]{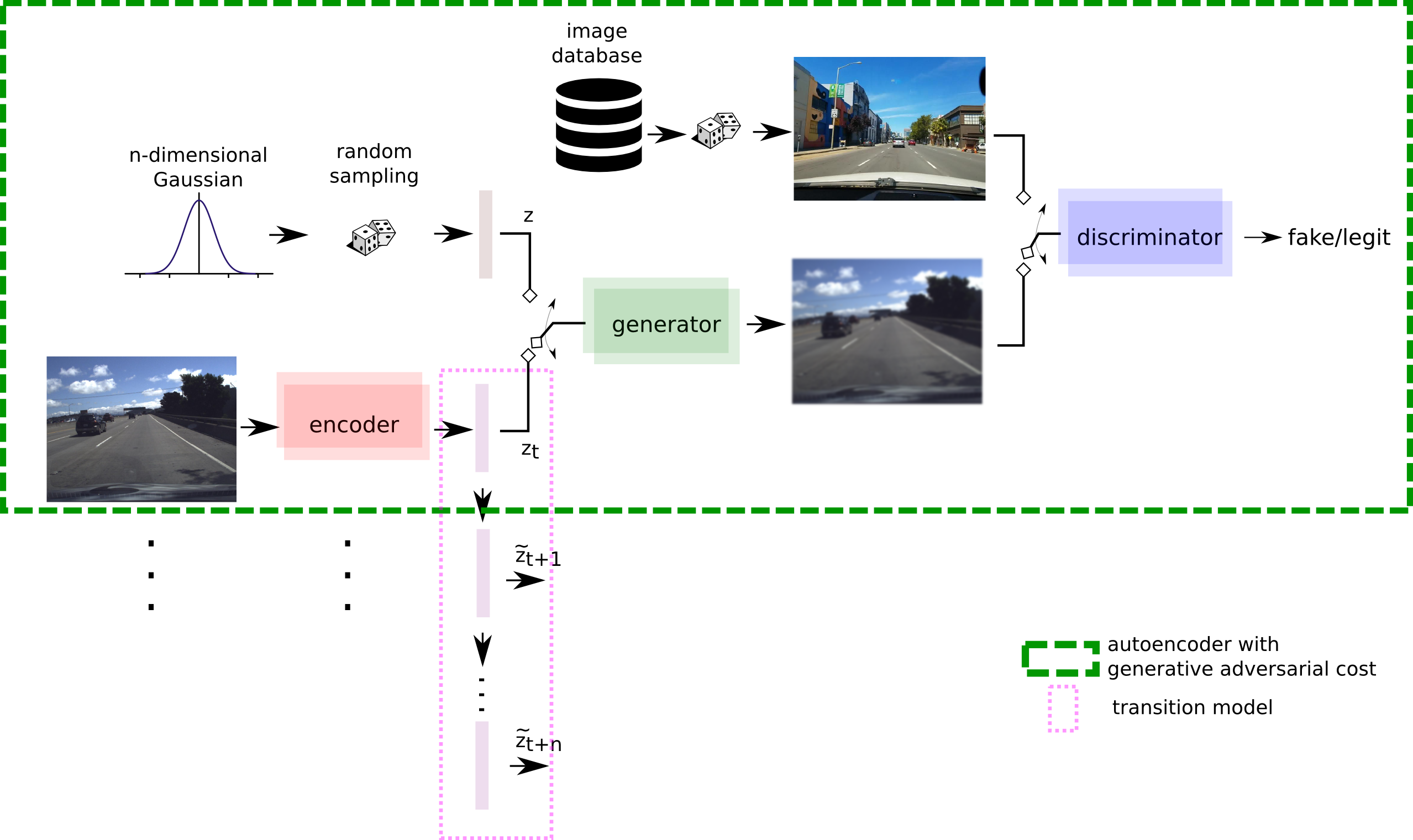}
 \caption{Driving simulator model: an autoencoder trained with generative adversarial costs coupled with a recurrent neural network transition model}
  \label{fig:aemodel}

\end{figure}

Due to the problem complexity we decided to learn video prediction with separable networks. We leave investigations of the end-to-end
approach for future work. The proposed architecture is based on two models:
an autoenecoder for dimensionality reduction and an action conditioned RNN for learning the transitions. Our complete model is depicted in
Fig. \ref{fig:aemodel}.

\subsection{Autoencoder}
\label{sub:ae}
\begin{figure}[t]
\centering
 \includegraphics[scale=1.0]{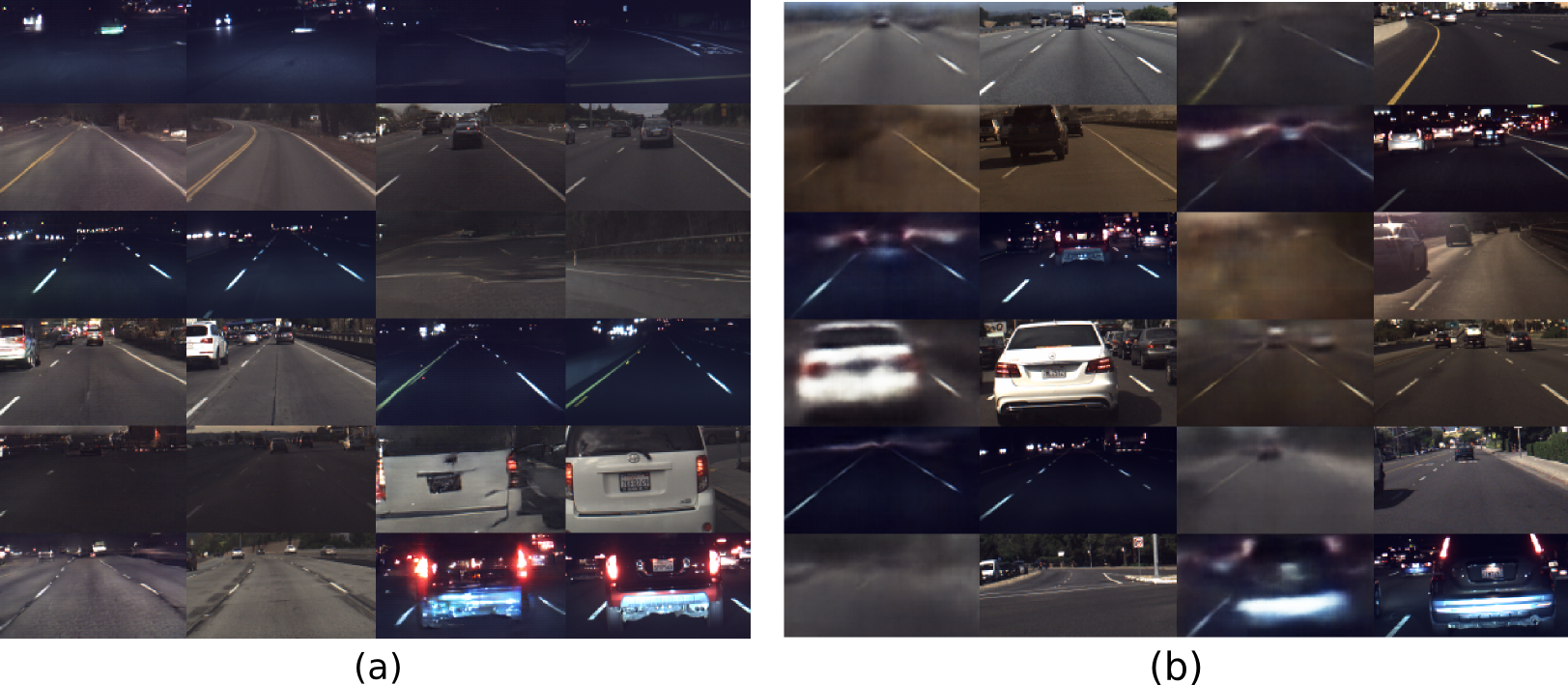}
 \caption{Samples using similar fully convolutional autoencoders. Odd columns show decoded images, even columns show target images.
 Models were trained using (a) generative adversarial networks cost function (b) mean square error. Both models have MSE in the order of $10^{-2}$ and PSNR in the order of $10$.}
   \label{fig:ganvsvae}

 \end{figure}

To learn the data embedding we chose a model that enforced the probability distribution of the latent space to be Gaussian. Specially to avoid
discontinuous regions of low probability inside a hypersphere centered in the origin. Such discontinuities could make it harder to
learn a continuous transition model in the latent space. Variational autoencoders \cite{VAE} and related work \cite{itlae} \cite{aae} had
recent sucess learning generative models with Gaussian priors in the latent and original data space. Unfortunately, the Gaussian assumption in the original
data space does not hold for natural images and VAE predictions usually look blurry (see Fig. \ref{fig:ganvsvae}).
On the other hand, generative adversarial networks (GAN) \cite{gan} and related work \cite{lapgan} \cite{dcgan} learn the generative model
cost function alongside the generator. They do so by alternating the training of generative and discriminator networks.
The generative model transforms samples from the latent space distribution into data from a desired dataset. The discriminator network
tries to tell samples from the desired dataset from images sampled by the generator. The generator is trained to "fool" the discriminator,
thus the discriminator can be considered a learned cost function for the generator.

For our purposes we need to learn not only the generator from the latent to the road images space. But also an encoder from road images
back to the latent space. We need to combine VAE and GANs. Intuitively, one could simply combine the VAE approach with a learned cost function.
In the literature Donahue et. al \cite{bigan} proposed bidirectional GANs that learns
both generation and encoding as bijective transformations. Lamb et. al. \cite{disgan} proposed discriminative generative networks that uses distances between
features of a previously trained classifier as part of the cost function. Finally, Larsen et. al. \cite{aegan} proposed to train a VAE alongside a GAN,
where the encoder optimizes both the Gaussian prior in the latent space and the similarity between features extracted by the GAN discriminator network.
The generator receives as inputs random samples from the latent space distribution and outputs of the encoder network. The generator is optimized to fool
the discriminator and minimize the similarity between original and decoded images. The discriminator is still only trained to tell fake from legit images.

We used Larsen et. al. \cite{aegan} approach to train our autoencoder. In Fig. \ref{fig:aemodel} we show the schematic diagram of this model as part of our architecture.
According to \cite{aegan} the encoder ($Enc$), generator ($Gen$) and discriminator ($Dis$) networks are optimized to minimize the following cost function:
\begin{equation}
 \label{eq:cost}
 \mathcal{L} = \mathcal{L}_{prior} + \mathcal{L}_{llike}^{Dis_l} + \mathcal{L}_{GAN}.
\end{equation}

In (\ref{eq:cost}) $\mathcal{L}_{prior} = D_{KL}(q(z|x)||p(z))$ is the Kullback-Liebler divergence between the distribution of the encoder outputs, $q(z|x)$,
and a prior distribution, $p(z)$.
This is the same VAE regularizer. Here the prior distribution is $p(z)$ is a Gaussian $\mathcal{N}(0, 1)$ and we use the \textit{reparametrization trick} \cite{VAE}
to optimize that regularizer. Thus, during training we have $z = \mu + \epsilon \sigma$ and at test time $z = \mu$, where $\mu$ and $\sigma$ are outputs
of the encoder network and $\epsilon$ is a Gaussian random vector with the same number of dimensions as $\mu$ and $\sigma$.

$\mathcal{L}_{llike}^{Dis_l}$ is an error calculated using $Dis_l$, the hidden activations of the $l$-th layer of the discriminator network, $Dis$.
The activations are calculated using a legit image $x$ and its corresponding encoded-decoded version $Gen(Dis(x))$. Assuming $y_l = Dis_l(x)$ and
$\tilde{y}_l = Dis_l(Gen(Enc(x)))$, we have $\mathcal{L}_{llike}^{Dis_l} = \mathbb{E}\left[ (y_l - \tilde{y}_l)^2 \right]$. During training, $Dis$
is considered constant with respect to this cost to avoid trivial solutions.

Finally, $\mathcal{L}_{GAN}$ is the generative adversarial network cost \cite{gan}. That cost function represents the \textit{game} between $Gen$ and $Dis$.
When training $Dis$, both $Enc$ and $Gen$ are kept fixed and we have
\begin{equation}
 \label{eq:disgan}
 \mathcal{L}_{GAN}^{Dis} = log(Dis(x)) + log(1 - Dis(Gen(u))) + log(1 - Dis(Gen(Enc(x))),
\end{equation}

where $u \sim \mathcal{N}(0, 1)$ is another random variable. The first r.h.t of (\ref{eq:disgan}) represents the log-likelihood of $Dis$ recognizing legit samples
and the other two terms represents its log-likelihood to tell fake samples generated from both random vectors $u$ or codes $z=Enc(x)$.
When training $Gen$, we keep both $Dis$ and $Enc$ fixed and we have
\begin{equation}
 \label{eq:gengan}
 \mathcal{L}_{GAN}^{Gen} = log(Dis(Gen(u))) + log(Dis(Gen(Enc(x))),
\end{equation}

which accounts for $Gen$ being able to fool $Dis$. Following \cite{aegan}, the gradient of $\mathcal{L}_{GAN}$ w.r.t $Enc$ is considered zero during training.

We trained this autoencoder for 200 epochs. Each epoch consisted of $10000$ gradient updates with a batch size of $64$.
Batches were sampled randomly from
driving data as described in the previous section. We used Adam for optimization \cite{adam}. The autoencoder architecture followed Radford et. al. \cite{dcgan},
with the generator made of 4 deconvolutional layers each one followed by batch normalization and leaky-ReLU activations. The discriminator and encoder
consisted of convolutional layers where each but the first layer was followed by batch normalization. The activation function was ReLU. $Dis_l$ is the output
of the 3rd convolutional layer of the decoder, before batch normalization and ReLU are applied. The output size of the
discriminator network is 1 and its cost function was binary cross-entropy. The output size of the encoder network is 2048. This compact representation is almost 16 times
smallet than the original data dimensionality. For network sizes and further details,
please check Fig. \ref{fig:aemodel} and the companion code of this paper. Sample encoded-decoded and target images are shown in Fig. \ref{fig:ganvsvae}.

After training the autoencoder, we fix all its weights and use $Enc$ as preprocessing step for training the transition model. We describe the transition model
in the next section.

\subsection{Transition model}
\label{sub:rnn}
After training the autoencoder described above we obtain the transformed dataset applying $Enc: x_t \mapsto z_t$. We train $RNN : z_t, h_t, c_t \mapsto z_{t+1}$ to
represent the transitions in the code space:

\begin{equation}
 \label{eq:rnn}
 \begin{split}
   h_{t+1} & = tanh\left( W h_t + V z_t + U c_t \right), \\
   \tilde{z}_{t+1} & = A h_{t+1}
 \end{split}
\end{equation}

where $W, V, U, A$ are trainable weights, $h_t$ is the hidden state of the RNN and $c_t$ are the concatenated control speed and angle signal.
We leave LSTMs, GRU and multiplicative iterations between $c_t$ and $z_t$ for future investigations.
The cost function for optimizing the trainable weights is simply the mean square error:

\begin{equation}
 \label{eq:rnn_mse}
 \mathcal{L}_{RNN} = \mathbb{E}\left[ (z_{t+1} - \tilde{z}_{t+1})^2 \right].
\end{equation}

It is easy to see that (\ref{eq:rnn_mse}) is optimal because we impose a Gaussian constraint $\mathcal{L}_{prior}$ in the distribution of the codes $z$ when training
the autoencoder. In other words, MSE equals up to a constant factor the log-likelihood of a normally distributed random variable.
Given a predicted code $\tilde{z}_{t+1}$ we can estimate future frames as $\tilde{x}_{t+1} = Gen(\tilde{z}_{t+1})$.

We trained a transition model with sequences of length of 15 frames. We used teacher forcing in the first 5 frames and fed the outputs back as new inputs in the remaining
10 frames. In other words, $z_1, ..., z_5$
were calculated using $Enc(x_t)$ and fed as input. We fed the outputs $\tilde{z}_{6}, ..., \tilde{z}_{15}$ back as input. In the RNN literature feeding outputs back
as input is informally referred as RNN \textit{hallucination}.
It is important to note that we consider the
gradient of the outputs equal to zero when they are fed back as inputs to avoid trivial solutions.

\begin{figure}[t]
\centering
 \includegraphics[scale=0.3]{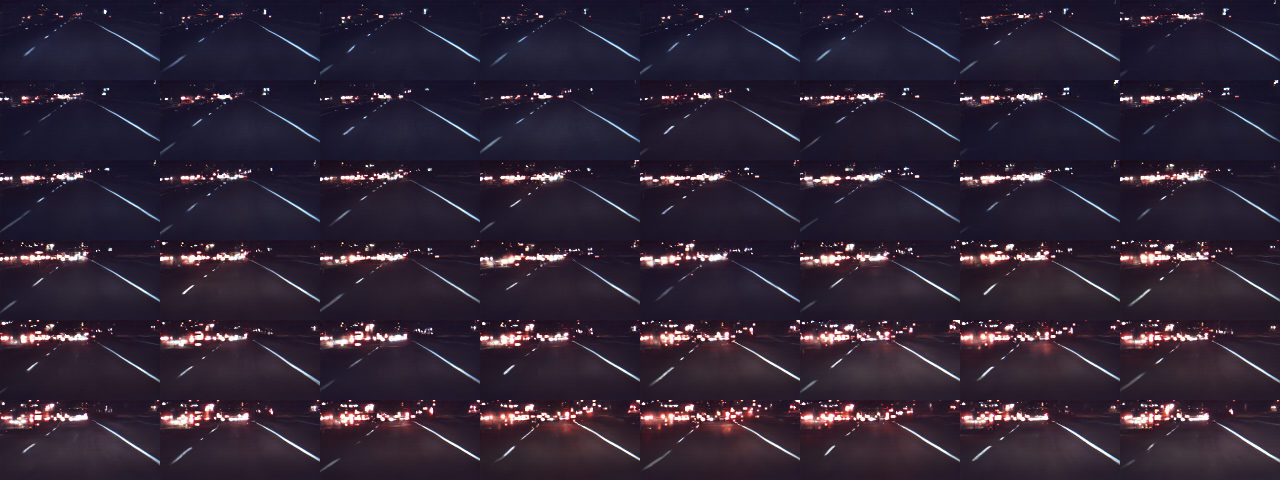}
 \caption{Samples generated by letting the transition model \textit{hallucinate} and decoding $\tilde{z}$ using $Gen$. Note that $Gen$ was not optimized to make these samples
 realistic, which support our assumption that the
 transition model did not leave the code space.}
   \label{fig:rnn_samples}

\end{figure}

\section{Results}
\label{sec:exp}
We spent most of the effort of this research investigating autoencoding architectures that could preserve the road texture. As mentioned before we studied different
cost functions. Although results were similar in terms of MSE, a learned cost function using generative adversarial networks had the most visually appealing results.
In Fig. \ref{fig:ganvsvae} we show
decoded images with models trained using two different cost functions. As expected, MSE based neural networks generate blurred images. The main issues
for our purposes is that blurring connects the lane marks into a single long lane. Also blurring reconstruction does not preserve leading car edges.
This is not desired for several reasons, the main one is that
it gets rid of one of the main clues used for visual odometry and leading car distance estimation. On the other hand MSE learns to generate curved lane
markings faster than the adversarial based model.
It is possible that learning to encode pixels with the car steering angle information should avoid this problem. We leave this investigation for future work.

Once we obtained a good autoencoder, we trained the transition model. Results of predicted frames are shown in Fig. \ref{fig:rnn_samples} .
We trained the transition model in 5Hz videos. The learned transition model keeps the road structure even after 100 frames.
When sampling from the transition model with different seed frames
we observed simulations of common driving events
such as passing lanes, approaching leading cars and leading cars moving away. The model failed to simulate curves. When we initialized the transition model with frames
of driving on a curve, the transition model quickly straightens the lanes and goes back to simulate \textit{going forward}.
Nevertheless, it is promissing that we could learn transitions for video generation without explicitly optimizing a cost function in the pixel space.
Also, we believe that even more realistic simulations will be possible with
more powerful transition models (deep RNNs, LSTM, GRU) and contextual encoding (sensor fusion of pixels plus steering angle and speed).
The dataset we are releasing with this paper has all the sensors necessary to experiment with this approach.

\section{Conclusions}
\label{sec:conclusions}

This paper presented \texttt{comma.ai} initial research on learning a driving simulator. We investigated video prediction models based on autoencoders and RNNs.
Instead of learning everything end-to-end
here we first trained the autoencoder with generative adversarial network based cost functions to generate realistic looking images of the road. We followed up
training an RNN transition model in the embedded space. Results from both autoencoder and
transition model look realistic but more research is needed to successfully simulate all the relevant events for driving. To stimulate further research on
this subject we are also releasing a driving dataset with video and several sensors such as car speed, steering angle, etc. We also release the code to train the networks
we investigated here.


\end{document}